\documentclass[
 aip,
 amsmath,amssymb,
 reprint,
]{revtex4-1}

\usepackage{graphicx}
\usepackage{dcolumn}
\usepackage{bm}

\usepackage[utf8]{inputenc}
\usepackage[T1]{fontenc}
\usepackage{mathptmx}
\usepackage{etoolbox}
\usepackage{color}
\usepackage{amsmath}
\usepackage{hyperref}

\newcommand{\D}{\mathrm{d}}

\makeatletter
\def\@email#1#2{%
 \endgroup
 \patchcmd{\titleblock@produce}
  {\frontmatter@RRAPformat}
  {\frontmatter@RRAPformat{\produce@RRAP{*#1\href{mailto:#2}{#2}}}\frontmatter@RRAPformat}
  {}{}
}%
\makeatother
\begin{document}

\preprint{AIP/123-QED}

\title{Learning Continuous Chaotic Attractors with a Reservoir Computer}
\author{Lindsay M. Smith}
\affiliation{Department of Physics \& Astronomy, University of Pennsylvania, Philadelphia, PA 19104 USA}
 
\author{Jason Z. Kim}%
\affiliation{Department of Bioengineering, University of Pennsylvania, Philadelphia, PA 19104 USA}

\author{Zhixin Lu}
\affiliation{Department of Bioengineering, University of Pennsylvania, Philadelphia, PA 19104 USA}

\author{Dani S. Bassett}
\email{dsb@seas.upenn.edu}
\affiliation{Department of Physics \& Astronomy, University of Pennsylvania, Philadelphia, PA 19104 USA}
\affiliation{Department of Bioengineering, University of Pennsylvania, Philadelphia, PA 19104 USA}
\affiliation{Department of Electrical \& Systems Engineering, University of Pennsylvania, Philadelphia, PA 19104 USA}
\affiliation{Department of Psychiatry, University of Pennsylvania, Philadelphia, PA 19104 USA}
\affiliation{Department of Neurology, University of Pennsylvania, Philadelphia, PA 19104 USA}
\affiliation{Santa Fe Institute, Santa Fe, NM 87501 USA}

\date{\today}

\begin{abstract}
Neural systems are well known for their ability to learn and store information as memories. Even more impressive is their ability to \textit{abstract} these memories to create complex internal representations, enabling advanced functions such as the spatial manipulation of mental representations. While recurrent neural networks (RNNs) are capable of representing complex information, the exact mechanisms of how dynamical neural systems perform abstraction are still not well-understood, thereby hindering the development of more advanced functions. Here, we train a 1000-neuron RNN --- a reservoir computer (RC) --- to abstract a continuous dynamical attractor memory from isolated examples of dynamical attractor memories. Further, we explain the abstraction mechanism with new theory. By training the RC on isolated and shifted examples of either stable limit cycles or chaotic Lorenz attractors, the RC learns a continuum of attractors, as quantified by an extra Lyapunov exponent equal to zero. We propose a theoretical mechanism of this abstraction by combining ideas from differentiable generalized synchronization and feedback dynamics. Our results quantify abstraction in simple neural systems, enabling us to design artificial RNNs for abstraction, and leading us towards a neural basis of abstraction.
\end{abstract}

\maketitle

\begin{quotation}
Neural systems learn and store information as memories, and can even create abstract representations from these memories, such as how the human brain can change the pitch of a song or predict different trajectories of a moving object. Because we do not know how neurons work together to generate abstractions, we are unable to optimize artificial neural networks for abstraction, or directly measure abstraction in biological neural networks. Our goal is to provide a theory for how a simple neural network learns to abstract information from its inputs. We demonstrate that abstraction is possible using a simple neural network, and that abstraction can be quantified and measured using existing tools. Further, we provide a new mathematical mechanism for abstraction in artificial neural networks, allowing for future research applications in neuroscience and machine learning.
\end{quotation}

\section{Introduction}
Biological and artificial neural networks have the ability to make generalizations from only a few examples \cite{Zhang2011,Faulkner2008,Dunn2012,Craik2006,Tacchetti2018,Moser2008,Ifft2013}. For instance, both types of networks demonstrate object invariance: the ability to recognize an object even after it has undergone translation or transformation \cite{guyonneau2006animals,zou2017learning}. What is surprising about this invariance is not that neural systems can map a set of inputs to the same output. Rather, what is surprising is that they can first sustain internal representations of objects, and then abstract these representations to include translations and transformations. Hence, beyond simply memorizing static, discrete examples \cite{hopfield1982neural}, neural systems have the ability to abstract their memories along a continuum of information by observing isolated examples \cite{seung1998learning}. However, the precise mechanisms of such abstraction remain unknown, limiting the principled design and training of neural systems.

To make matters worse, much of the information represented by neural networks is not static, but \textit{dynamic}. As a biological example, a songbird's representation of song is inherently time-varying, and can be continuously sped up and slowed down through external perturbations \cite{Fee2010}. In artificial networks, recurrent neural networks (RNNs) can store a history of temporal information such as language \cite{mikolov2010recurrent}, dynamical trajectories \cite{Jaeger2010,Sussillo2009}, and climate \cite{nadiga2021reservoir} to more accurately classify and predict future events. To harness the power of RNNs for processing temporal information, efforts have focused on developing powerful training algorithms such as backpropagation through time (BPTT) \cite{lillicrap2019backpropagation} and neural architectures such as long short-term memory (LSTM) networks \cite{hochreiter1997long}, alongside physical realizations in neuromorphic computing chips \cite{furber2016large}. Unfortunately, the dramatic increase in computational capability is accompanied by a similarly dramatic increase in the difficulty of understanding such systems, severely limiting their designability and generalizability beyond specific datasets.
 
To better understand the mechanisms behind neural representations of temporal information, the field has turned to dynamical systems. Starting with theories of synchronization between coupled dynamical systems \cite{davison2016synchronization,pecora1990synchronization}, theories of generalized synchronization \cite{Rulkov1995} and invertible generalized synchronization \cite{lu2020invertible} provide intuition and conditions for when a neural network uniquely represents the temporal trajectory of its inputs, and when this representation can recover the original inputs to recurrently store them as memories \cite{lu2018attractor}. These theories hinge on important ideas and tools such as delay embedding \cite{garcia2005multivariate}, Lyapunov exponents \cite{dawson1994obstructions}, and dimensionality \cite{young1982dimension,frederickson1983liapunov}, which quantify crucial properties of time-varying representations.  However, it is not yet known precisely how neural systems \textit{abstract} such time-varying representations. Accordingly, the field is limited in its understanding of abstraction and meta-learning in existing neural systems\cite{Kumar2020metalearning,Schweighofer2003metalearning,Santiago2004context,Feldkamp1997adaptive}, and restricted in its ability to design neural systems for abstraction.

Here, we address this knowledge gap by providing a mechanism for the abstraction of time-varying attractor memories in a reservoir computer (RC) \cite{lukovsevivcius2009reservoir}. First, we demonstrate that a neural network can observe low dimensional inputs and create higher dimensional abstractions, thereby learning a continuum of representations from a few examples. Then, we develop new theory to explain the mechanism of this abstraction by extending prior work \cite{kim2021teaching}: we explicitly write the differential response of the RC to a differential change in the input, thereby giving a quantitative form to ideas of differentiable generalized synchronization \cite{hunt1997differentiable}. We quantify this abstraction by demonstrating that successful abstraction is driven by the acquisition of an additional 0 Lyapunov exponent in the RC's dynamics, and study the role of the RC's spectral radius and time constant on its ability to abstract dynamics. These results enable the development of more interpretable and designable methods in machine learning, and provide a quantitative hypothesis and measure of abstraction from neural dynamics.

\section{Mathematical Framework}
\begin{figure}[h]
\centering
\includegraphics[width=\columnwidth]{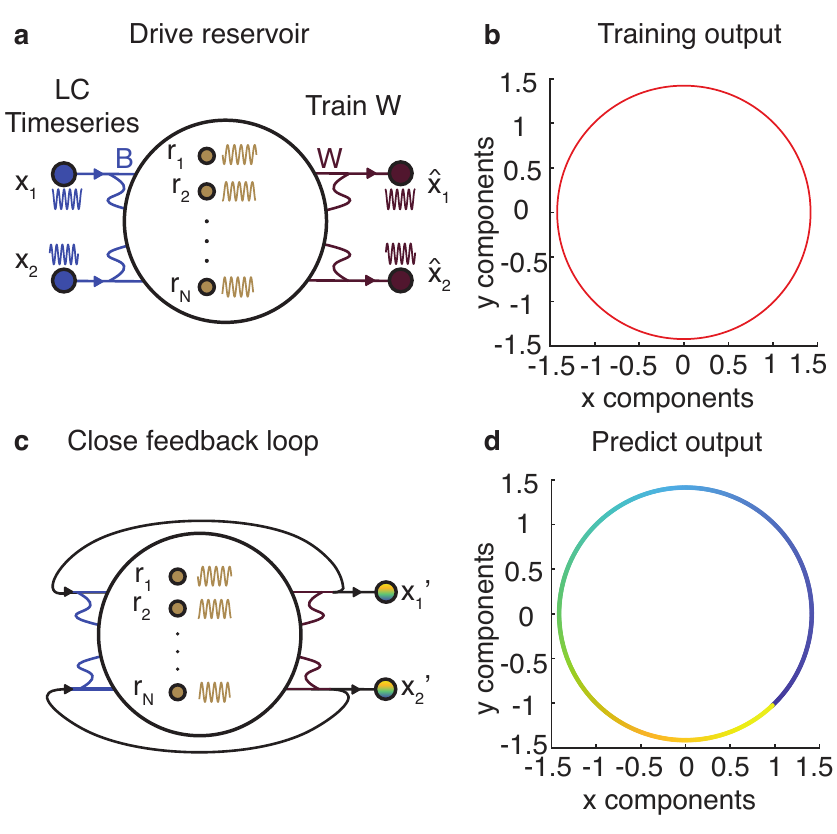}
\caption{Schematic of a reservoir computer learning a limit cycle memory. (a) Time series of a limit cycle that drives the RNN reservoir to the state of the limit cycle. Weighted sums of the reservoir states are trained to reproduce the original time series (b) by creating the $W$ matrix. (c) The reservoir uses the weighted sums in $W$ to evolve, closing the feedback loop in the RC. (d) The RC now evolves autonomously along a trajectory that closely follows the expected dynamics of the original limit cycle. Here, color represents time.}
\label{fig:schematic}
\end{figure}
To study the ability of neural networks to process and represent time-varying information as memories, we use a simple nonlinear dynamical system from reservoir computing:

\begin{equation}
\frac{1}{\gamma}\dot{\bm{r}}(t) = -\bm{r}(t) + \bm{g}(A\bm{r}(t) + B\bm{x}(t) + \bm{d})
\label{eq:open_loop}.
\end{equation}
Here, $\bm{r}(t) \in \mathbb{R}^{N \times 1}$ is a vector that represents the state of the $N$ reservoir neurons, $\bm{x}(t) \in \mathbb{R}^{k \times 1}$ is the vector of $k$ inputs into the reservoir, $\bm{d} \in \mathbb{R}^{N \times 1}$ is a constant vector of bias terms, $A \in \mathbb{R}^{N \times N}$ is the matrix of connections between neurons, $B \in \mathbb{R}^{N \times k}$ is the matrix of weights mapping inputs to neurons, $\bm{g}$ is a sigmoidal function that we take to be the hyperbolic tangent $\tanh$, and $\gamma$ is a time constant. 

Throughout the results, we use an $N=1000$-neuron network, such that $A\in\mathbb{R}^{1000 \times 1000}$. We set $A$ to be $2\%$ sparse, where each non-zero entry of $A$ is a random number from -1 to 1, and then scale $A$ such that the absolute value of the largest eigenvalue is $\rho$, the spectral radius of the network. In general, each entry of $B$ was drawn randomly from -1 to 1, and multiplied by a scalar coefficient set to 0.1; one analysis that stands as an exception is the parameter sweep in Subsection \ref{paramsweep}, where the scalar coefficient was varied systematically. Each entry of the bias term $\bm{d}$ was drawn randomly from -1 to 1, and multiplied by a bias amplification constant, which was set to 10 in all cases except in the parameter sweep in Subsection \ref{paramsweep}.

To study the ability of the reservoir to form representations and abstractions of temporal data, we must define the data to be learned. Following prior work in teaching reservoirs to represent temporal information \cite{kim2021teaching,lu2018attractor,Jaeger2010,Sussillo2009}, we will use dynamical attractors as the memories. The first memory that we use is a stable limit cycle that evolves according to
\begin{equation}
\begin{split}
\dot{x}_1(t) &= 10x_1(t)(2-x_1^2(t)-x_2^2(t))-10x_2(t)\\
\dot{x}_2(t) &= 10x_1(t)(2-x_1^2(t)-x_2^2(t))+10x_1(t)
\end{split}
\label{eq:limit_cycle}.
\end{equation}
To test the reservoir's ability to learn and abstract more complex memories, the second memory that we use is the chaotic Lorenz attractor \cite{Lorenz1963} that evolves according to
\begin{equation}
\begin{split}
\dot{x}_1(t) &= -10(x_1(t)-x_2(t))\\
\dot{x}_2(t) &= 28x_1(t)-x_2(t)-x_1(t)x_3(t)\\
\dot{x}_3(t) &= -8/3x_3(t)+x_1(t)x_2(t)
\end{split}
\label{eq:lorenz}.
\end{equation}
By driving the reservoir in Eq.~\ref{eq:open_loop} with the time series generated from either the stable limit cycle in Eq.~\ref{eq:limit_cycle} or the chaotic Lorenz system in Eq.~\ref{eq:lorenz}, the response of the reservoir neurons is given by $\bm{r}(t)$. In our experiments, we drive the reservoir and evolve the input memory for 50 seconds to create a transient phase which we discard, allowing the RC and the input memory to evolve far enough away from the randomly chosen initial conditions. Then, we drive the reservoir and the input memory together for 100 seconds to create a learning phase. Because we use a time step of $\D t = 0.001$, this process creates a learning phase time series of 100,000 points.

To store the attractor time series as memories, prior work in reservoir computing has demonstrated that it is sufficient to first train an output matrix $W$ that maps reservoir states $\bm{r}(t)$ to copy the input $\bm{x}(t)$ according to
\begin{equation}
\begin{split}
\min_W \|W\bm{r}(t) - \bm{x}(t)\|_2^2,
\label{eq:training}
\end{split}
\end{equation}
and then perform feedback by replacing the inputs $\bm{x}(t)$ with the output of the reservoir, $W\bm{r}(t)$. This feedback generates a new system that evolves autonomously according to
\begin{equation}
\frac{1}{\gamma}\dot{\bm{r}}(t) = -\bm{r}(t) + \tanh((A+BW)\bm{r}(t) + \bm{d}).
\label{eq:closed_loop}
\end{equation}
We evolve this new system for 500 seconds to create a prediction phase.

As a demonstration of this process, we show a schematic of the reservoir being driven by the stable limit cycle input (Fig.~\ref{fig:schematic}a, blue), thereby generating the reservoir time series (Fig.~\ref{fig:schematic}a, gold), which is subsequently used to train a matrix $W$ such that $W\bm{r}(t)$ copies the input (Fig.~\ref{fig:schematic}a, red). The training input, $\bm{x}(t)$ (Fig.~\ref{fig:schematic}b, blue), and the training output, $W\bm{r}(t)$ (Fig.~\ref{fig:schematic}b, red), are plotted together and are indistinguishable. After the training, we perform feedback by replacing the reservoir inputs, $\bm{x}(t)$, with the outputs, $W\bm{r}(t)$ (Fig.~\ref{fig:schematic}c), and observe that the output of the autonomous reservoir remains as a limit cycle (Fig.~\ref{fig:schematic}d). Can this simple process be used not only to store memories, but also to abstract memories? If so, by what mechanism?

In what follows, we answer these questions by extending the framework to multiple isolated inputs. Specifically, rather than use only one attractor time series $\bm{x}_0(t)$, we will use a finite number of translated attractor time series 
\begin{equation}
\begin{split}
\bm{x}_c(t) = \bm{x}_0(t) + c\bm{a}
\end{split}
\label{eq:shift}
\end{equation}
for $c \subset \mathbb{Z}$, where $\bm{a}$ is a constant vector. We will use these time series to drive the reservoir to generate a finite number of neural responses $\bm{r}_c(t)$. By concatenating all of the inputs and reservoir states along the time dimension into a single time series, $\bm{x}(t)$ and $\bm{r}(t)$, respectively, we train an output matrix $W$ according to Eq.~\ref{eq:training} that maps all of the reservoir states to all of the translated inputs. Finally, using $W$, we perform feedback according to Eq.~\ref{eq:closed_loop}.

\section{Differential learning drives abstraction}
\begin{figure}[h]
\centering
\includegraphics[width=\columnwidth]{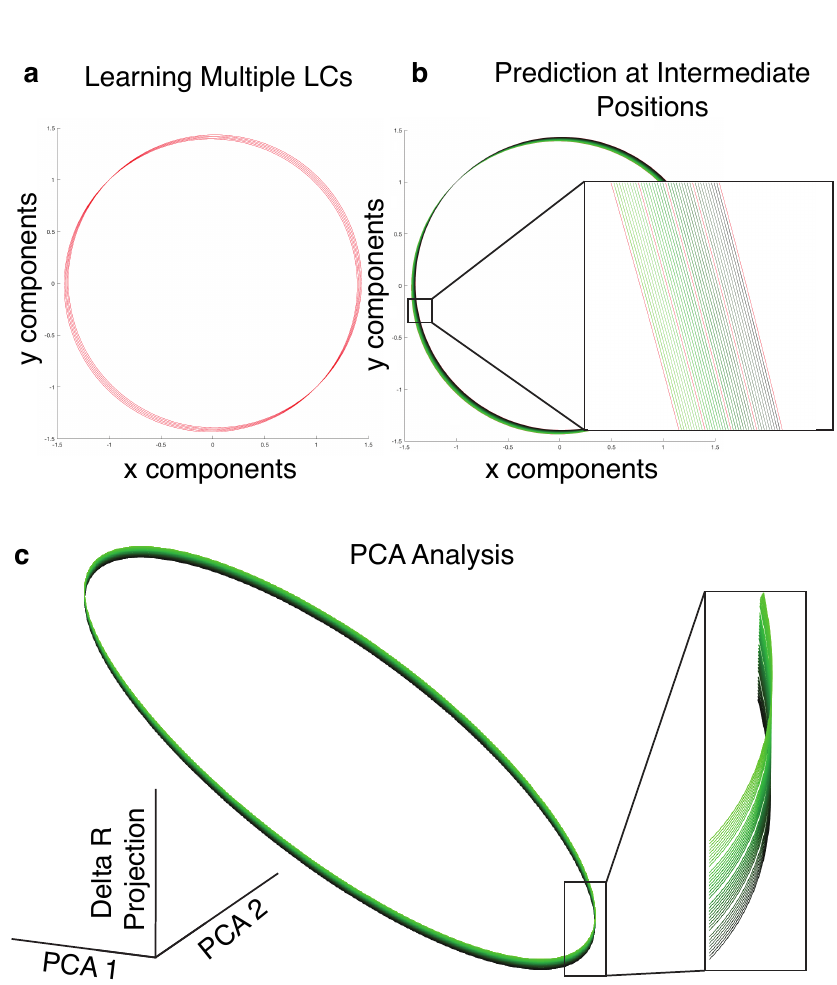}
\caption{Successful abstraction in learning a continuous limit cycle memory. (a) 5 shifted limit cycles are learned by the reservoir as 5 isolated examples. (b) 2D plot of the predicted output of the autonomous reservoir whose initial state has been prepared between the 5 training examples. The shift magnitude, or the distance of each translation, of the initial state is colored from green to black. (c) To visualize the abstraction that occurred in an additional dimension along the direction of the translation, we show a 3D plot of the predicted reservoir time series projected onto the $\Delta \bm{r}$ vector in Eq.~\ref{eq:deltar}, and the first two principal components after removing the $\Delta r$ projection. The cutout highlights the "height" of the the continuous attractor, formed from the abstraction along the $\Delta \bm{r}$ axis.}
\label{fig:fig2}
\end{figure}
To teach the reservoir to generate higher-dimensional representations of isolated inputs, we train it to copy translations of an attractor memory. First, we consider the time series of a stable limit cycle generated by Eq.~\ref{eq:limit_cycle}, $\bm{x}_0(t)$, and we create shifted time series, $\bm{x}_c(t)$, for $c \in \{-2,-1,0,1,2\}$ according to Eq.~\ref{eq:shift} (Fig.~\ref{fig:fig2}a). We then use these time series to drive the reservoir according to Eq.~\ref{eq:open_loop} to generate the reservoir time series $\bm{r}_c(t)$ for $c \in \{-2,-1,0,1,2\}$, concatenate the time series into $\bm{x}(t)$ and $\bm{r}(t)$, respectively, and train the output matrix $W$ according to Eq.~\ref{eq:training} to generate the autonomous feedback reservoir that evolves according to Eq.~\ref{eq:closed_loop}.

To test whether the reservoir has learned a higher-dimensional continuum of limit cycles \textit{versus} the five isolated examples, we evolve the autonomous reservoir at intermediate values of the translation variable $c$. Specifically, we first prepare the reservoir state by driving the non-autonomous reservoir in Eq.~\ref{eq:open_loop} with limit cycles at intermediate translations (i.e. $\bm{x}_c(t)$ for $c \in \{-1.9, -1.8, -1.7, \dotsm, 1.7, 1.8, 1.9\}$) for 50 seconds until any transient dynamics from the initial reservoir state have decayed, thereby generating a set of final reservoir states $\bm{r}_c(t=50)$. We then use these final reservoir states as the initial state for the autonomous feedback reservoir in Eq.~\ref{eq:closed_loop}. Finally, we evolve the autonomous reservoir and plot the outputs in green in Fig.~\ref{fig:fig2}b. As can be seen, the autonomous reservoir whose initial state has been prepared at intermediary shifts, translations in position, continues to evolve about a stable limit cycle at that shift.

\subsection{Differential mechanism of learning}
Now that we have numerically demonstrated the higher-dimensional abstraction of lower-dimensional attractors, we will uncover the underlying theoretical mechanism first by studying the response of the reservoir to different inputs, and then by studying the consequence of the training process.

First, we compute perturbations of the reservoir state, $\D\bm{r}(t)$, in response to perturbations of the input, $\D\bm{x}(t)$, by linearizing the dynamics about the trajectories $\bm{r}_0(t)$ and $\bm{x}_0(t)$ to yield
\begin{equation}
\begin{split}
\D\dot{\bm{r}}(t) = \underbrace{\gamma(\D\bm{g}(t) \circ A - I)}_{A^*(t)}\D\bm{r} + \underbrace{\gamma\D\bm{g}(t)\circ B}_{B^*(t)}\D\bm{x}(t),
\end{split}
\label{eq:linearized}
\end{equation}
where $\D\bm{g}(t)$ is the derivative of $\tanh(\cdot)$ evaluated at $A\bm{r}_0(t) + B\bm{x}_0(t)$, and $\circ$ is the element-wise product of the $i$-th element of $\D\bm{g}(t)$ and the $i$-th row of either matrix $A$ or $B$. We are guaranteed by differentiable generalized synchronization \cite{hunt1997differentiable} that if $\D\bm{x}(t)$ is infinitesimal and constant, then $\D\bm{r}(t)$ is also infinitesimal and evolves according to Eq.~\ref{eq:linearized}. Fortuitously, the differential change $\D\bm{x}(t)$ is precisely infinitesimal and constant, and is given by the derivative of Eq.~\ref{eq:shift} to yield $\D\bm{x}(t) = \bm{a}\D c$. We substitute this derivative into Eq.~\ref{eq:linearized} to yield
\begin{equation}
\begin{split}
\D\dot{\bm{r}}(t) = A^*(t)\D\bm{r}(t) + B^*(t)\bm{a}\D c.
\end{split}
\label{eq:ltv}
\end{equation}
Crucially, this system is \textit{linear}, such that if a shift of $\bm{a}\D\bm{c}$ yields a perturbed reservoir trajectory of $\D\bm{r}(t)$, then a shift of $2\bm{a}\D\bm{c}$ yields a perturbed reservoir trajectory of $2\D\bm{r}(t)$. Hence, we can already begin to see the mechanism of abstraction: any scalar multiple of the differential input, $\bm{a}\D c$, yields a scalar multiple of the trajectory $\D\bm{r}(t)$ as a valid perturbed trajectory.

To complete the abstraction mechanism, we note that the trained output matrix precisely learns the inverse map: if Eq.~\ref{eq:ltv} maps scalar multiples of $\bm{a}\D c$ to scalar multiples of $\D\bm{r}(t)$, then the trained output matrix $W$ maps scalar multiples of $\D\bm{r}(t)$ back to $\bm{a}\D c$. To learn this inverse map, notice that our 5 training examples are spaced closely together (Fig.~\ref{fig:fig2}a), which allows the trained output matrix $W$ to map \textit{differential} changes in $\bm{r}(t)$ to \textit{differential} changes in $\bm{x}(t)$. Hence, not only does $W\bm{r}_c(t) \approx \bm{x}_c(t)$, but $W$ also learns
\begin{equation}
\begin{split}
W\D\bm{r}(t) \approx \D\bm{x}(t) = \bm{a}\D c.
\end{split}
\label{eq:differential}
\end{equation}
The consequence of this differential learning is seen in the evolution of the perturbation $\D\bm{r}(t)$ of the autonomous feedback reservoir by substituting Eq.~\ref{eq:differential} into Eq.~\ref{eq:ltv} to obtain
\begin{equation}
\begin{split}
\D\dot{\bm{r}}(t) = A^*(t)\D\bm{r}(t) + B^*(t)W\D\bm{r}(t).
\end{split}
\label{eq:ltv_feedback}
\end{equation}
If the training examples are close enough to learn the differential relation in Eq.~\ref{eq:differential}, then any perturbed trajectory, $\D\bm{r}(t)$, generated by Eq.~\ref{eq:ltv} is a valid trajectory in the feedback system to linear order. Further, any \textit{scalar multiple} of $\D\bm{r}(t)$ is also a valid perturbed trajectory in the feedback system. 

Hence, by training the output matrix to copy nearby examples---thereby learning the differential relation between $\D\bm{r}(t)$ and $\D\bm{x}(t)$---we encode scalar multiples of $\D\bm{r}(t)$ as a linear subspace of valid perturbation trajectories. It is precisely this encoded subspace of valid perturbation trajectories that we call the \textit{higher-dimensional abstraction} of the lower-dimensional input; in addition to the 2-dimensional limit cycle input, the reservoir encodes the subspace comprising scalar multiples of the perturbation trajectory as a third dimension. To visually represent this third dimension, we take the average of the perturbation vector across time as
\begin{equation}
\begin{split}
\Delta \bm{r} = \frac{1}{T} \int_0^T \D\bm{r}(t)\D t,
\end{split}
\label{eq:deltar}
\end{equation}
and project all of the autonomous reservoir trajectories along this vector. We remove this projection from the first two principal components of the autonomous reservoir trajectories, and then we plot the same projection against these two modified principal components, shown in Fig.~\ref{fig:fig2}c. As can be seen, the shift in the limit cycle is encoded along the $\Delta \bm{r}$ direction. Graphically and numerically, we have confirmed our theoretical mechanism of abstraction using a continuous limit cycle memory.

\section{Abstraction as the acquisition of a Lyapunov Exponent equal to zero}
\begin{figure}[h]
\centering
\includegraphics{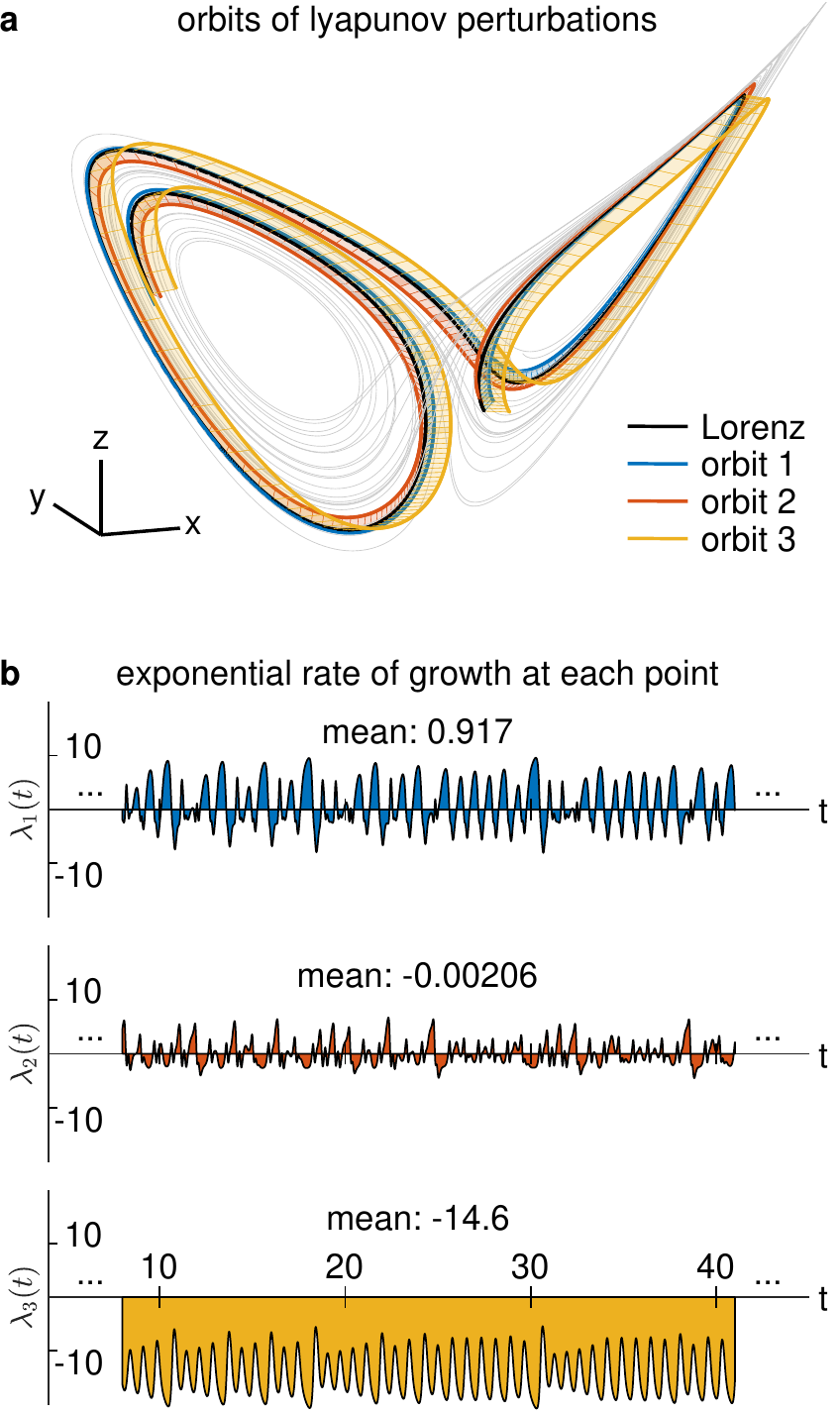}
\caption{Obtaining a Lyapunov spectrum from a Lorenz attractor. (a) A 3D plot of the Lyapunov perturbation orbits (colored), which check the stability of a trajectory, about the Lorenz attractor (black, gray), obtained by evolving the orbits about the Jacobian of the Lorenz system evaluated at each point, followed by an orthonormalization. (b) A plot of the three Lyapunov exponents over $\approx 30$ seconds for the Lorenz system, whose average over $T=100$ seconds yields the estimated Lyapunov exponents.}
\label{fig:fig3}
\end{figure}
Now that we have a mechanism of abstraction, we seek a simple method to quantify this abstraction in higher-dimensional systems that do not permit an intuitive graphical representation (Fig.~\ref{fig:fig2}c). In the chaotic Lorenz system with a fractal orbit (Fig.~\ref{fig:fig3}a), it can be difficult to visually determine whether the prediction output is part of a given input example or in between two input examples. Hence, we would like some measure of the presence of perturbations along a trajectory, $\D\bm{r}(t)$, that neither grow nor shrink along the direction of linearly scaled perturbations. If these perturbations neither grow nor shrink, they represent a stable trajectory that does not collapse into another trajectory or devolves into chaos. 

To measure this abstraction, we compute the Lyapunov spectrum of the RC. Conceptually, the Lyapunov spectrum measures the stability of different trajectories along an attractor. It is computed by first generating an orbit along the attractor of a $k$-dimensional dynamical system, $\bm{x}(t)$, then by evaluating the Jacobian at every point along the orbit, $J(\bm{x}(t))$, and finally by evolving orbits of infinitesimal perturbations, $\bm{p}_i(t)$, along the time-varying Jacobian as
\begin{equation}
\begin{split}
\dot{\bm{p}}_i(t) = J(\bm{x}(t)) \bm{p}_i(t).
\end{split}
\label{eq:lyapunov_state}
\end{equation}
Along these orbits, the direction and magnitude of $\bm{p}_i(t)$ will change based on the linearly stable and unstable directions of the Jacobian $J(\bm{x}(t))$. To capture these changes along orthogonal directions, after each time step of evolution along Eq.~\ref{eq:lyapunov_state}, we order the perturbation vectors into a matrix $[\bm{p}_1(t), \bm{p}_2(t), \dotsm, \bm{p}_k(t)]$, and perform a Gram-Schmidt orthonormalization to obtain an orthonormal basis of perturbation vectors $\tilde{\bm{p}}_1(t), \tilde{\bm{p}}_2(t), \dotsm, \tilde{\bm{p}}_k(t)$. In this way, $\tilde{\bm{p}}_1(t)$ eventually points along the least stable direction, $\tilde{\bm{p}}_2(t)$ along the second least stable direction, and $\tilde{\bm{p}}_k(t)$ along the most stable direction. The evolution of the three perturbation vectors of the Lorenz system are shown in Fig.~\ref{fig:fig3}a.

To calculate the Lyapunov spectrum, we compute the projection of each normalized perturbation vector along the Jacobian as the Lyapunov exponent (LE) over time\cite{balcerzak2018fastest},
\begin{equation}
\begin{split}
\lambda_i(t) = (J(\bm{x}(t)) \tilde{\bm{p}_i}(t)) \cdot \tilde{\bm{p}_i}(t),
\end{split}
\label{eq:lyapunov_exponent}
\end{equation}
and the final LE is given by the time average (Fig.~\ref{fig:fig3}b). Every continuous-time dynamical system with bounded, non-fixed-point dynamics has at least one zero Lyapunov exponent corresponding to a perturbation that neither grows nor shrinks on average (Fig.~\ref{fig:fig3}b, red). In a chaotic system like the Lorenz, there is also a positive LE corresponding to an orthogonal perturbation that grows on average (Fig.~\ref{fig:fig3}b, blue). Finally, a negative LE corresponds to an orthogonal perturbation that decays on average (Fig.~\ref{fig:fig3}b, yellow). As can be seen in the plot of trajectories, the orbit of the negative LE is directed transverse to the plane that roughly defines the ``wings'' of the attractor, such that any deviation from the plane of the wings quickly collapses back onto the wings (Fig.~\ref{fig:fig3}a).

Using the Lyapunov spectrum, we hypothesize that the reservoir's abstraction of an attractor memory will appear as an additional LE equal to zero. This is because through the training of nearby examples, the reservoir acquires the perturbation direction $\D\bm{r}(t)$ that neither grows nor decays on average, as all scalar multiples of $\D\bm{r}(t)$ are valid perturbation trajectories to linear order according to Eq.~\ref{eq:ltv_feedback}. Hence, the acquisition of such a perturbation direction that neither grows nor decays should present itself as an additional LE equal to zero. 

\subsection{Abstraction depends on spectral radius and time constant} \label{paramsweep}
\begin{figure}[h]
\centering
\includegraphics[width=\columnwidth]{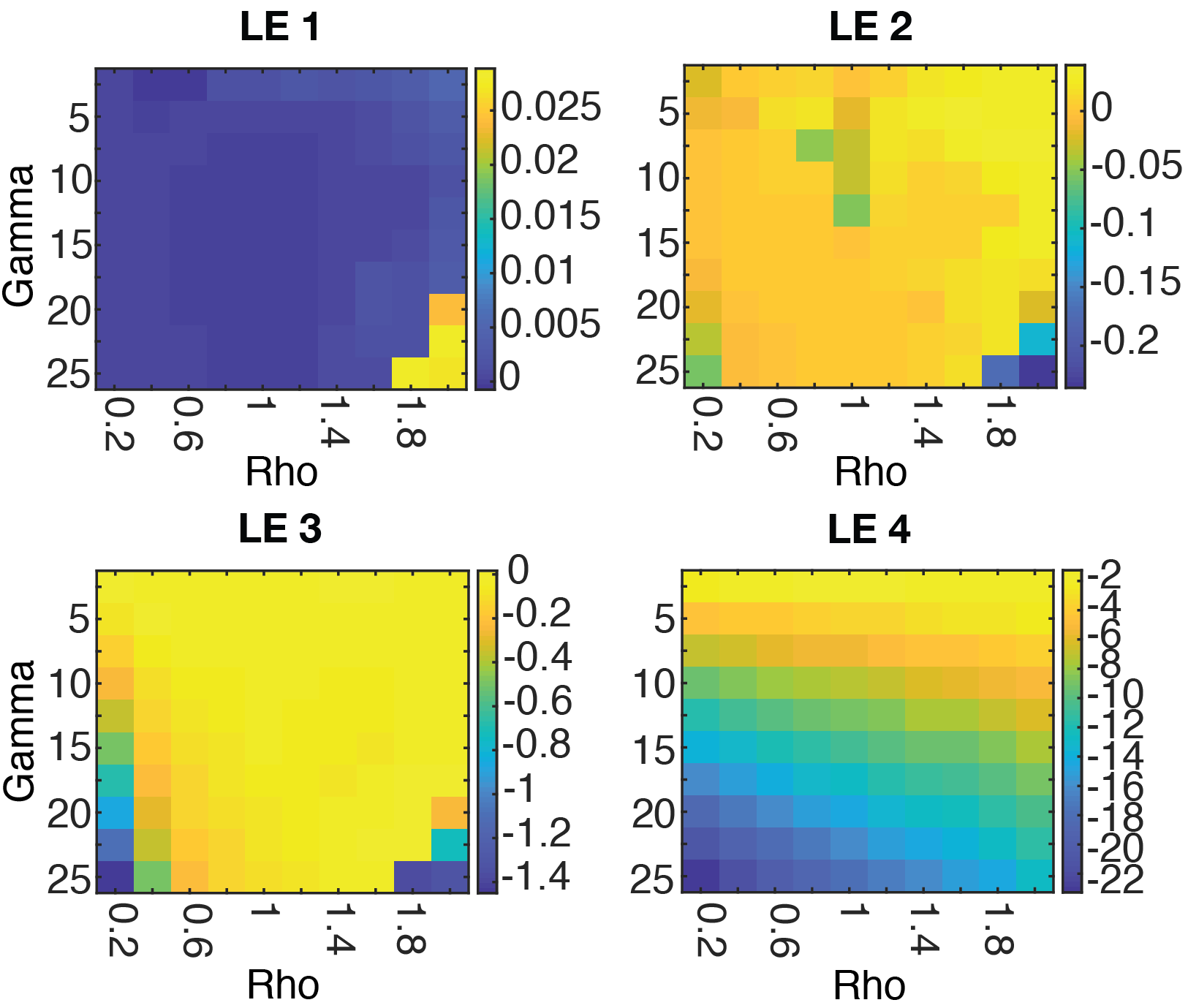}
\caption{LEs of an RC after learning a continuous limit cycle. Heat maps of the first 4 LEs of the RC with different spectral radii ($\rho$) on the $x$-axis and time constants ($\gamma$) on the $y$-axis.}
\label{fig:fig4}
\end{figure}

With this mechanism of abstraction in RNNs, we provide a concrete implementation of our theory and study its limits. Our RNNs in Eq.~\ref{eq:open_loop} and in Eq.~\ref{eq:closed_loop} depend on several parameter regimes; the spectral radius (given by $\rho$), the time constant $\gamma$, the bias term $\bm{d}$, the weighting of the input matrix $B$, and the number and spacing of the training examples all impact whether abstraction can successfully occur. We quantify the effect of varying these parameters on the RC's ability to abstract different inputs \textit{via} the Lyapunov spectrum analysis.

We focus on the parameters that determine the internal dynamics of the RC: $\gamma$ and $\rho$. The RC is a carefully balanced system whose internal speed is set by $\gamma$. If $\gamma$ is too small, the system is too slow to react to the inputs. And conversely, if $\gamma$ is too large, the system responds too quickly to retain a history of the input. Thus, we hypothesize that an intermediate $\gamma$ will yield optimal abstraction, and that the optimal range of $\gamma$ will vary depending upon the time scale of the input. Similar to the time constant, the spectral radius is known to impact the success of the learning as it controls the excitability of the RNN \cite{Sussillo2009}. For abstraction to succeed, the RNN needs an intermediate $\gamma$ and $\rho$ to learn the input signals with an excitability and reaction speed suited for the input attractor memory.

To find the ideal parameter regime for abstracting a limit cycle attractor memory, we performed a parameter sweep on $\gamma$ from 2.5 to 25.0 in increments of 2.5, and on $\rho$ from 0.2 to 2.0 in increments of 0.2. All other parameters in the closed and open loop reservoir equations were held constant. To measure the success of the abstraction, we calculated the first 4 LEs of the RC, looking for values of $\lambda_1$ and $\lambda_2$ equal to 0, and values of $\lambda_3$ and $\lambda_4$ that are negative. For this continuous limit cycle memory, we found that the best parameter regime was $0.4<\rho<1.0$ and $20<\gamma<25$ (Fig.~\ref{fig:fig4}). Then, we tested the weighting of the input matrix, $B$, while holding $\rho$, $\gamma$, and all other parameters constant. We found that an optimal scaling of $B$ is between 0.001 and 0.1. We performed a similar test for the bias term, $\bm{d}$, resulting in an optimal scaling between 1.0 and 20.0. These parameter ranges demonstrate that a careful balance of $\rho$ and $\gamma$, along with $B$ and $\bm{d}$, is necessary to successfully achieve abstraction. More generally, our approach to defining these parameter ranges provides a principled method for future RNNs to learn different attractor memories.

\subsection{Abstraction of chaotic memories}
\begin{figure}[h]
\centering
\includegraphics[width=\columnwidth]{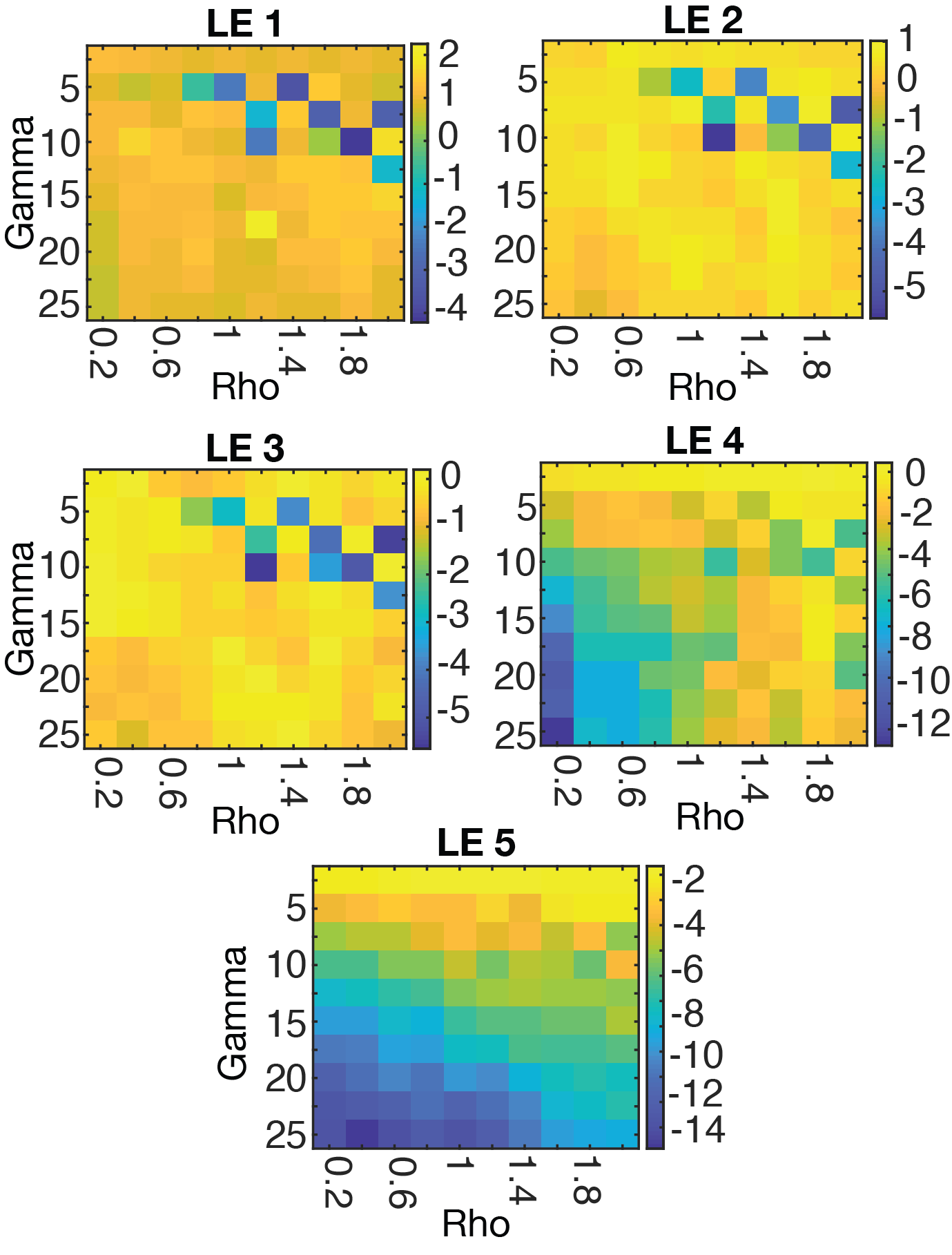}
\caption{LEs of a RC after learning a continuous Lorenz attractor. Heat maps of the first 5 LEs of the RC with different spectral radii ($\rho$) on the $x$-axis and time constants ($\gamma$) on the $y$-axis.}
\label{fig:fig5}
\end{figure}

While limit cycles provide an intuitive conceptual demonstration of abstraction, real neural networks such as the human brain learn more complex memories that involve a larger number of parameters, including natural and chaotic attractors such as weather phenomena\cite{Lorenz1963} or diffusion\cite{pathak2017using}. Chaotic attractors pose a more complex memory for the reservoir to learn, so it is nontrival to show that the reservoir is able to abstract from several chaotic attractors to learn one single continuous chaotic attractor. By again analyzing the Lyapunov spectrum, we can quantify successful abstraction. As seen in Figure ~\ref{fig:fig3}, a chaotic dynamical system is characterized by positive Lyapunov exponents. In the case of the Lorenz attractor, the first Lyapunov exponent is positive ($\approx0.9$), the second is equal to zero, and the third is negative ($\approx-14.6$). Hence, when the RC learns a single Lorenz attractor, the first LE is positive, the second LE is zero, and the rest of the spectrum is increasingly negative. In the case of the successful and continuous abstraction of the Lorenz attractor, we expect to see that the first LE is positive, followed by not one, but \textit{two} LEs equal to zero, followed by increasingly negative LEs.

To test this acquisition of an LE equal to zero, we trained the RNN to learn multiple chaotic attractor memories, focusing on the Lorenz attractor. To find the ideal parameter regime for learning a continuous Lorenz attractor memory from many discrete examples, we again performed a parameter sweep on $\gamma$ from 2.5 to 25.0 in increments of 2.5, and on $\rho$ from 0.2 to 2.0 in increments of 0.2. All other parameters in the closed and open loop reservoir equations were held constant. To measure the success of abstraction, we calculated the first 5 LEs of the RC, looking for the values of $\lambda_1$ to be positive,  values of $\lambda_2$ and $\lambda_3$ to equal zero, and the values of $\lambda_4$ and $\lambda_5$ to be increasingly negative. For this continuous Lorenz attractor memory, we found that the best parameter regime was using 0.6-1.2 for $\rho$ and a $\gamma$ of 25, as seen in Fig.~\ref{fig:fig5}. Hence, we demonstrate that in addition to simple limit cycle attractors, RNNs can succesfully abstract much more complex and unstable chaotic attractor memories, demonstrating the generalizability of our theory.

\section{Discussion}
Reservoir computing has been gaining substantial traction, and significant advances have been made in many domains of application. Among them include numerical advances in adaptive rules for training reservoirs using evolutionary algorithms \cite{ferreira2009genetic,ferreira2013approach} and neurobiologically-inspired spike-time-dependent-plasticity \cite{paugam2008delay}. In tandem, physical implementations of reservoir computing in photonic \cite{katumba2017multiple,salehi2014optical,rohm2019reservoir}, memristive\cite{merkel2014memristive}, and neuromorphic \cite{donati2018processing} systems provide low-power alternatives to traditional computing hardware. Each application is accompanied by its own unique set of theoretical considerations and limitations \cite{koster2020limitations}, thereby emphasizing the need for the underlying analytical mechanisms to make meaningful generalizations across such a wide range of systems.

In this work, we provide such a mechanism for the abstraction of a continuum of attractor memories from discrete examples, and put forth the acquisition of an additional zero Lyapunov exponent as a quantitative measure of success. Moreover, the method can be applied to any learning of chaotic attractor memories due to the generality of the differential mechanism of learning we uncover. While our investigation simplifies the complexity of the network used and the memories learned, we show that the underlying mechanism of abstraction remains the same as we increase the complexity of the memory learned (e.g., discrete to continuous, and non-chaotic to chaotic).

Our work motivates several new avenues of inquiry. First, it would be of interest to examine the theoretical and numerical mechanism for abstracting more complex transformations. Second, it would be of interest to embark on a systematic study of the spacing between the discrete examples that is necessary to learn a differential attractor \textit{versus} discrete attractors, and the phase transition of abstraction. Third and finally, ongoing and future efforts could seek to determine the role of noise in both the RNN and input dynamics for abstracting high-dimensional continuous attractors from scattered low-dimensional and discrete attractors. Because different RNNs are better suited to learn and abstract different inputs, we expect that this work will shed light on studies that reveal how one can design specialized RNNs for better abstraction on particular dynamical attractors.

\section{Conclusion}
Here we show that an RNN can successfully learn time-varying attractor memories. We demonstrate this process with both limit cycle and Lorenz attractor inputs. We then show the RNN several discrete examples of these attractors, translated from each other by a small distance. We find that the neural network is able to abstract to a higher dimension and learn a continuous attractor memory that connects all of the discrete examples together. This process of abstraction can be quantified by the acquisition of an additional exponent equal to zero in the Lyapunov spectrum of the RC's dynamics. Our discovery has important implications for future improvements in the algorithms and methods used in machine learning, due specifically to the understanding gained from using this simpler model. More broadly, our findings provide new hypotheses regarding how humans construct abstractions from real-world inputs to their neural networks.

\section{Acknowledgments}
LMS acknowledges support from the University Scholars Program at the University of Pennsylvania. JZK acknowledges support from the NIH T32-EB020087, PD: Felix W. Wehrli, and the National Science Foundation Graduate Research Fellowship No. DGE-1321851. DSB acknowledges support from the NSF through the University of Pennsylvania Materials Research Science and Engineering Center (MRSEC) DMR-1720530, as well as the Paul G. Allen Family Foundation, and a grant from the Army Research Office (W911NF-16-1-0474). The content is solely the responsibility of the authors and does not necessarily represent the official views of any of the funding agencies.

\section{Citation diversity Statement}
We would like to include a citation diversity statement following a recent proposal \cite{zurn2020citation}. Recent work in several fields of science has identified a bias in citation practices such that papers from women and other minority scholars are under-cited relative to the number of such papers in the field \cite{mitchell2013gendered,dion2018gendered,caplar2017quantitative, maliniak2013gender, Dworkin2020.01.03.894378, bertolero2021racial, wang2021gendered, chatterjee2021gender, fulvio2021imbalance}. Here we sought to proactively consider choosing references that reflect the diversity of the field in thought, form of contribution, gender, race, ethnicity, and other factors. First, we obtained the predicted gender of the first and last author of each reference by using databases that store the probability of a first name being carried by a woman \cite{Dworkin2020.01.03.894378,zhou_dale_2020_3672110}. By this measure (and excluding self-citations to the first and last authors of our current paper), our references contain 6.67\% woman(first)/woman(last), 16.4\% man/woman, 13.33\% woman/man, and 63.6\% man/man. This method is limited in that a) names, pronouns, and social media profiles used to construct the databases may not, in every case, be indicative of gender identity and b) it cannot account for intersex, non-binary, or transgender people. Second, we obtained predicted racial/ethnic category of the first and last author of each reference by databases that store the probability of a first and last name being carried by an author of color \cite{ambekar2009name, sood2018predicting}. By this measure (and excluding self-citations), our references contain 15.94\% author of color (first)/author of color(last), 20.06\% white author/author of color, 20.17\% author of color/white author, and 43.83\% white author/white author. This method is limited in that a) names and Florida Voter Data to make the predictions may not be indicative of racial/ethnic identity, and b) it cannot account for Indigenous and mixed-race authors, or those who may face differential biases due to the ambiguous racialization or ethnicization of their names.  We look forward to future work that could help us to better understand how to support equitable practices in science.

\section{References}
\bibliography{references2}
\end{document}